\documentclass[10pt,twocolumn,letterpaper]{article}

\usepackage[pagenumbers]{wacv} 

\definecolor{wacvblue}{rgb}{0.21,0.49,0.74}
\usepackage[pagebackref,breaklinks,colorlinks,allcolors=wacvblue]{hyperref}

\usepackage{multirow}
\usepackage{diagbox}
\usepackage[table]{xcolor}
\usepackage{graphicx}

\usepackage{booktabs}
\usepackage{colortbl}
\usepackage{amsmath}
\usepackage{makecell}
\usepackage[accsupp]{axessibility}  

\newcommand{\myparagraph}[1]{\vspace{1.5pt}\noindent{\bf #1}}
\newcommand{\acro}[0]{MM-TS\xspace}

\def\wacvPaperID{1920} 
\def\confName{WACV}
\def\confYear{2026}

\title{MM-TS: Multi-Modal Temperature and Margin Schedules for Contrastive Learning with Long-Tail Data}

\author{
Siarhei Sheludzko$^{1}$ \quad
Dhimitrios Duka$^{2}$ \quad
Bernt Schiele$^{2}$ \quad
Hilde Kuehne$^{3, 4}$ \quad
Anna Kukleva$^{2}$\\
\vspace{0.3em}
{\small $^{1}$University of Bonn \quad
$^{2}$MPI for Informatics, SIC \quad
$^{3}$Tuebingen AI Center/University of Tuebingen \quad 
$^{4}$MIT-IBM Watson AI Lab}
}

\begin{document}
\maketitle
\begin{abstract}

Contrastive learning has become a fundamental approach in both uni-modal and multi-modal frameworks. This learning paradigm pulls positive pairs of samples closer while pushing negatives apart. In the uni-modal setting (e.g., image-based learning), previous research has shown that the strength of these forces can be controlled through the temperature parameter.
In this work, we propose Multi-Modal Temperature and Margin Schedules (\acro), 
extending the concept of uni-modal temperature scheduling to multi-modal contrastive learning. Our method dynamically adjusts the temperature in the contrastive loss during training, modulating the attraction and repulsion forces in the multi-modal setting.
Additionally, recognizing that standard multi-modal datasets often follow imbalanced, long-tail distributions, we adapt the temperature based on the local distribution of each training sample. Specifically, samples from dense clusters are assigned a higher temperature to better preserve their semantic structure.
Furthermore, we demonstrate that temperature scheduling can be effectively integrated within a max-margin framework, thereby unifying the two predominant approaches in multi-modal contrastive learning: InfoNCE loss and max-margin objective.
We evaluate our approach on four widely used image- and video-language datasets, Flickr30K, MSCOCO, EPIC-KITCHENS-100, and YouCook2, and show that our dynamic temperature and margin schedules improve performance and lead to new state-of-the-art results in the field. \footnote{The code is publicly available at \url{https://github.com/SergShel/MM-TS}} 
\vspace{-4mm}
\end{abstract}    
\section{Introduction}
\label{sec:intro}

Contrastive learning has become one of the most important tools in self-supervised learning, allowing models to learn representations from single modalities, such as images~\cite{he2019momentum, chen2020simple, zhang2022dual}, as well as from multiple modalities, such as vision and language~\cite{radford2021learning,Cherti_2023_CVPR}. By aligning different modalities of the same instance in a joint embedding space, 
these models achieve state-of-the-art performance, e.g., in the context of zero-shot classification, image retreival~\cite{guo2024m2encoderadvancingbilingualimagetext,dehghani2023lit_scalingvisiontransformers22,pham2023combinedscalingzeroshottransfer} and video retrieval~\cite{wang2024vamos,zhao2023training,zhao2023learning}, but also serve as a representation backbone for autoregressive frameworks~\cite{liu2023improvedllava, liu2023llava, Maaz2023VideoChatGPT}.




Contrastive learning
aims to bring positive pairs closer together, such as different augmentations of the same image in the uni-modal case or image-caption pairs in the multi-modal case, while pushing apart all other samples, so called negatives.
In particular, in the multi-modal setting, contrastive learning aligns representations from two different modalities by minimizing the cosine distance between matching text-image pairs.
This process is controlled by the temperature parameter $\tau$ in InfoNCE-based losses or by the margin definition in the max-margin formulations, both of which control the strength of attraction and repulsion forces in the embedding space.

While most approaches simply fix the temperature parameter during training~\cite{he2020momentum, chen2020improved, chen2020simple} or consider it a constant hyperparameter, recent works also started to explore how to vary or adapt it to improve contrastive training itself~\cite{kukleva2023temperature, wang2021understanding, kim2025temperature}. 
%
Among them, temperature schedules~\cite{kukleva2023temperature} (TS) explore the effect of controlling temperature to improve contrastive self-supervised learning for imbalanced distributions. 
Low temperatures amplify pushing forces of negatives, leading to stronger discrimination between individual instances,
while 
a higher temperature exerts less force on negative samples thus enhancing grouping based on semantic structure.
By dynamically scheduling the temperature during training, the model continuously learns different semantic features: lower temperatures enhance instance discrimination, while higher temperatures lead to group-wise structure formation.
Training under this regime has shown to be valuable 
as this is especially important for single-modality imbalanced distributions~\cite{kukleva2023temperature}.


In this work we propose a 
\underline{M}utli-\underline{M}odal \underline{T}emperature and Margin \underline{S}chedules (\acro) 
framework for multi-modal contrastive learning on long-tail data. Our \acro leverages the idea of temperature schedules and adapts it for the multi-modal scenario
while at the same time including information about the local distribution of the multi-modal data. 
To this end, first, we employ a dynamic temperature scheduling for both modalities that follows a cosine schedule. Continuous changes of the embedding space allow the model to learn different semantic features and improve its alignment compared to a fixed temperature. 
Second, we leverage the availability of the two modalities and approximate the distribution of the visual training data by using the text modality.
We propose to control the amplitude of the temperature parameter based on the distributions of the training data. 
Namely, when the corresponding narration contains an object or a concept that frequently occurs in the dataset,
we assign it a higher value. 
This allows the creation of larger clusters in the representation space for those samples, so called group-wise discrimination effect, whereas pairs with rare and unique narrations would repel with a higher strength from all other pairs enforcing instance discrimination. 

Finally, we extend the temperature scheduling formulation from  the original InfoNCE loss function to the max-margin contrastive loss setting. We observe similar behavior that indicates the importance of the repulsion forces of negatives as the max-margin loss does not explicitly control strength of positive pulling.

We evaluate the influence of the proposed \acro framework for the case of long-tail image- and video-language learning, considering image-text settings by pretraining on CC3M with the zero-shot retrieval on Flickr30k and COCO, and video-text datasets such as the fine-grained long-tail EPIC-KITCHENS-100 dataset~\cite{damen2022rescaling}, and YouCook2 dataset~\cite{zhou2018towards} .
To this end, we extend one of the best performing contrastive methods of each framework and retrain it with the proposed \acro extensions. 



Our contributions can be summarized as following: 
\begin{itemize}
    \item We propose a new multi-modal  framework for contrastive learning on long-tail data that controls temperature parameters by combining cosine temperature schedule with individual adjustments for each sample based on the estimated distributions; 
    \item We generalize temperature scheduling beyond the conventional InfoNCE framework, extending it to the widely adopted max-margin loss. This extension is particularly impactful within communities such as egocentric video analysis, where max-margin loss remains a prevalent choice and datasets are long-tailed.
    \item We evaluate the proposed apporach across various naturally distributed long-tail image and video datasets such as CC3M, YouCook2 and EPIC-KITCHENS-100. 
\end{itemize}

\section{Related work}
\label{sec:related_work}




\myparagraph{Uni-modal Contrastive Learning} (CL) is a self-supervised learning approach that aims to learn semantically meaningful representations from a single modality, e.g. images, text, or audio, by leveraging an objective that maximizes the similarity between positive pairs,  usually constructed by augmenting the original data sample, and minimizes the similarity between negative pairs. 
Building on the importance of negative samples in contrastive learning \cite{NEURIPS2020_f7cade80, 2219be86fb9a441aac9d0e91ca3c782a, zhang2022dual, yeh2021decoupled}, 
MoCo~\cite{he2020momentum, chen2020improved} introduces CL
with momentum encoder and memory bank in the form of a FIFO queue, 
used to store generated features from the momentum encoder which are later used to construct negative pairs for CL. SimCLR~\cite{chen2020simple} refines MoCo’s approach by eliminating the momentum and dependency on the memory bank by increasing the batch size and introducing a projection head. 

\myparagraph{Multi-Modal Contrastive Learning.} The contrastive learning methods have also been used to learn semantically rich representations in the context of multimodal data. CLIP~\cite{radford2021learning} constructs a joint representation space by employing the InfoNCE objective in multi-modal embedding and learning an alignment for image-text pairs.
This simple yet effective approach enables zero-shot transfer to many tasks and datasets. 
Different methods have aimed to improve CLIP's performance by applying architectural changes \cite{lu2024improving}, refining the optimization process \cite{yuan2022provable}, or redefining the objective \cite{zhang2021temperature, cui2022democratizing, mu2021slip, li2021supervision, naeem2024silc}. SogCLR \cite{yuan2022provable} introduces the idea of a global contrastive objective that aims to contrast a positive pair against all negative pairs in the dataset. 
CyCLIP~\cite{goel2022cyclip} addresses the issue of inconsistent image and text representations in CLIP-like models by introducing an extension of the classical InfoNCE loss and proposing two additional objectives to ensure geometric consistency within and between modalities. 
SigCLIP \cite{zhai2023sigmoid} replaces the classical softmax-based InfoNCE objective with a pairwise sigmoid objective. The sigmoid objective is introduced for image-text pairs, thus eliminating the need for normalization across an entire mini-batch, resulting in less computation, higher efficiency, and scalability. 

 
\myparagraph{Temperature in Contrastive Learning.} Recent works \cite{wang2021understanding, qiu2024cool, kim2025temperature, li2023curriculum, wang2020contextual, zhang2021temperature, kukleva2023temperature} have shown that temperature $\tau$ has a great impact on CL, influencing both the uniformity and the alignment of the embedding space. \cite{wang2021understanding} showed that the contrastive objective is hardness-aware. At small temperatures, the objective penalizes the hard negatives more, compared to soft negatives. This behavior results in instance discrimination, making the embedding space more uniform at the cost of a less aligned embedding space. At high temperatures, the disparity in penalties between soft and hard negatives is reduced, therefore all negatives are penalized more uniformly. 
This encourages the formation of clusters and trades-off uniformity for alignment~\cite{kukleva2023temperature}. TempNet~\cite{qiu2024cool} proposed using an additional temperature prediction network to assign a unique temperature per sample. Even though the idea of using a network to predict a per-sample temperature was considered before \cite{li2023curriculum, wang2020contextual, zhang2021temperature}, TempNet introduces a novel design based on Distributionally Robust Optimization (DRO) and variational analysis. \cite{kim2025temperature} proposed a hyperparameter-free method that replaces the temperature on the CL with an inverse hyperbolic tangent function.
In contrast to the above-mentioned methods, our \acro method takes a much simpler approach. Instead of altering the objective, architecture, or the optimization process,
our method relies on a standard CLIP objective and modifies only the temperature parameter during the training process.




\myparagraph{Imbalanced Self-Supervised Learning} 
aims to tackle the challenging problem of learning semantically meaningful representations without labeled data under the constraints imposed by an imbalanced dataset. Previous works addressed this problem through e.g. resampling \cite{han2005borderline, drummond2003c4} or reweighing \cite{cui2019class, park2021influence}. 
Alternatively, \cite{chen2023area} argued that using the number of samples as a proxy to calculate the entity weight is not ideal and propose a distribution-based analysis which also accounts for relations between samples, whereas recent work on CL provides another approach to dealing with imbalanced data. \cite{kukleva2023temperature} shows that using a simple temperature scheduler in uni-modal CL can result in a more uniform embedding space and can address the problem of underrepresented classes. 
On one hand, under a low-temperature regime, the model shows a good instance-discrimination behavior, from which tail classes can benefit. On the other hand, under a high-temperature regime, the model "switches"  to cluster creation behavior, from which highly-represented classes can benefit. 
Moreover, recent studies, i.a. \cite{kang2020exploring, liu2021self_1, yang2020rethinking}, show that compared to fully supervised methods, self-supervised methods can learn more robust feature representations.
%
%
\begin{figure}
    \centering
    \includegraphics[width=\linewidth]{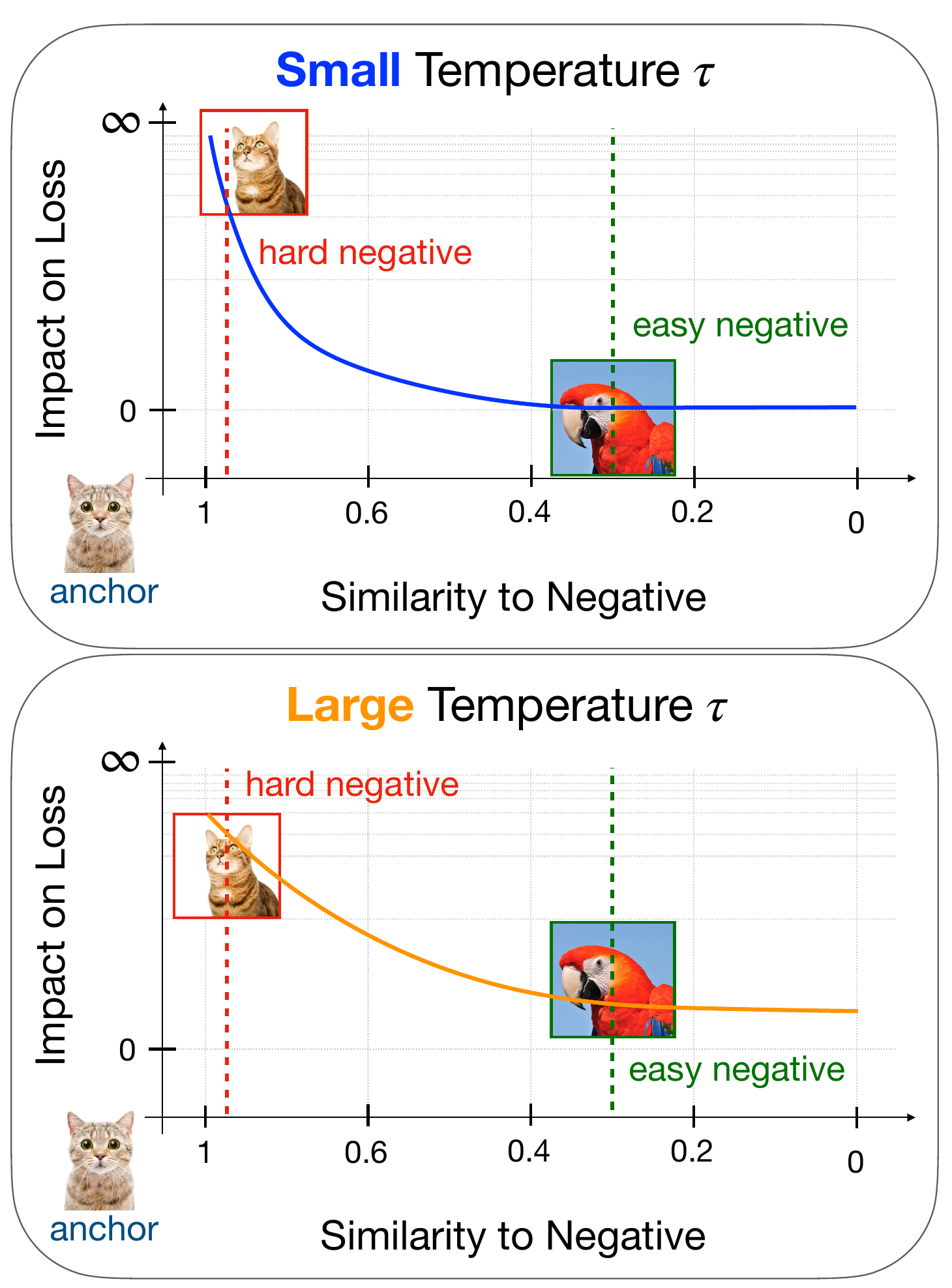}
    \caption{Individual impact of negatives to the loss depends on the temperature parameter. With the small temperature, only hard negatives impact the loss, whereas impact of easy negatives is negligible. With the large temperature impact of easy negatives to the loss increases.  }
    \label{fig:impact_tau}
    \vspace{1mm}
\end{figure}

\section{Method}
\label{sec:method}




In this section, we first review the principles of uni-modal contrastive learning. In particular, we discuss InfoNCE and Max-Margin losses and the impact of the temperature parameter ($\tau$), showing its important role in
controlling semantic relationships among data-points. Then, we revisit mutli-modal contrastive loss. 
Finally, inspired by uni-modal finding on contrastive learning,   we propose our method that refines the structure of the multi-modal embeddings and improve representations for 
long-tailed data. 

\subsection{Background}

Contrastive learning (CL)  is one of the most important frameworks to learn expressive representations that are used in various downstream tasks~\cite{chen2020improved, chen2020simple}.
Given a set of inputs $\{x_1,...,x_N\}$ and visual encoder $f_v$, the similarity score between the two inputs can be computed as:
\begin{equation}
    s_{ij} = f_v(x_i)f_v(x_j)^T.
    \label{eq:similarity1}
\end{equation}
\myparagraph{InfoNCE Loss}~\cite{oord2018representation} is a commonly used objective, formulated as:

%
%
\begin{equation}
   \mathcal{L}_{\text{InfoNCE}}(s) = -\frac{1}{N} \sum_{i=1}^{N} \log \frac{\exp(s_{ii} / \tau)}{ \sum_{j=1}^{N} \exp(s_{ij} / \tau)},
  \label{eq:info-nce}
\end{equation}
where $s_{ii}$ represents cosine similarity between representations of input $x_i$ and its random augmentation $\hat{x}_i$ (\eg different crops of the same image), and $s_{ij}$ is similarity between representations of $x_i$ and $x_j$ inputs. Further, we refer to $x_i$ as an anchor, $\hat{x}_i$ as a positive, and $x_j$ for all $j \neq i$ as negatives.  $\tau$ denotes the temperature parameter. Such formulation of the loss encourages pulling together the positive samples, whereas negative samples are pushed away.

\begin{figure*}
  \centering
  \includegraphics[width=1\linewidth]{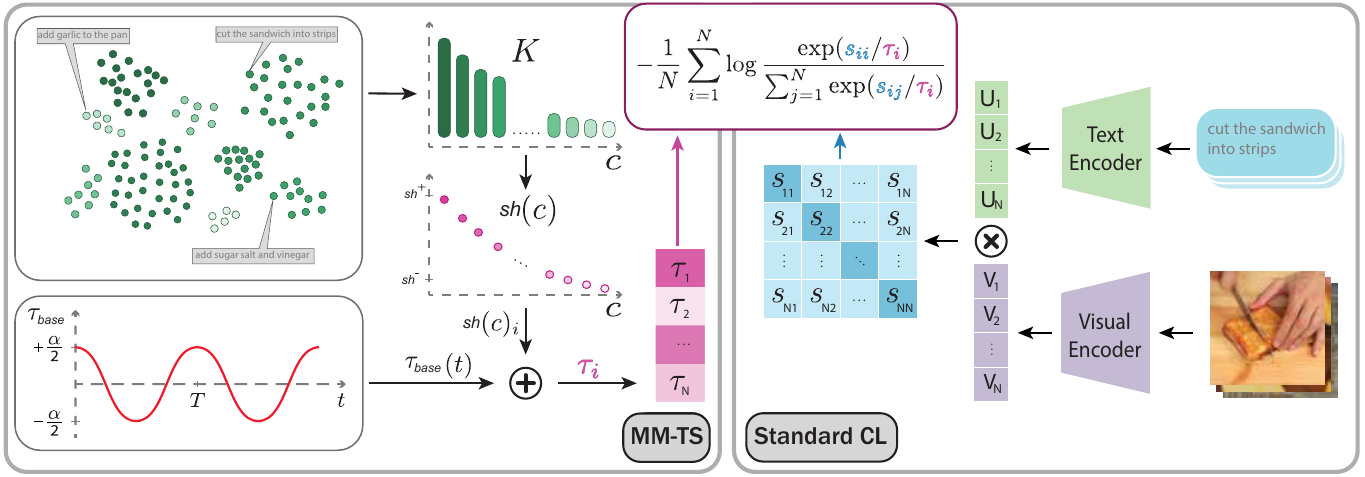}
  \caption{Visualization of our \acro approach for computing the InfoNCE loss. First, we cluster text annotations to estimate the distribution of the training data. For each cluster, we assign the respective cluster-based shift ($sh(c)$), where to the larger cluster we assign a larger shift. The base temperature $\tau_{base}$ follows the cosine schedule. Next, we adjust the temperature for each individual sample based on the estimate cluster-based shifts resulting in $\tau_i$. The updated individual temperature $\tau_i$ is used in a standard InfoNCE loss. 
  }
  \label{fig:pipeline_main}
  \vspace{1mm}
\end{figure*}

\myparagraph{Max-Margin Loss}~\cite{schroff2015facenet} is
one of the most popular contrastive loss alternatives to InfoNCE loss. The objective of the max-margin loss is to maximize the distance between positive and negative pairs and is formulated as: 
\begin{equation}
    \mathcal{L}_{\text{max-margin}} = \max \left( 0, \, s_{ij} - s_{ii} + m \right)
    \label{eq:max-margin}
\end{equation}
with 
$m$ being the margin which defines the minimum distance required between the positive and negative pairs.

\myparagraph{Influence of the temperature parameter $\tau$} As shown i.a. in~\cite{kukleva2023temperature, wang2021understanding}, the influence of $\tau$ can be important for the learning of representations. Specifically in \cite{kukleva2023temperature}, the authors discuss impact of the temperature parameter on the learning of the semantic structure in the representation space.
A smaller $\tau$ amplifies the impact of the nearest negatives (hard negatives) while the impact of more distant (easy) negatives is negligible, see \cref{fig:impact_tau} top. Therefore, 
only the closest, hardest negatives contribute to the loss. 
Compared to that, with a larger $\tau$ individual contributions of each negatives are on the same order of magnitude, see \cref{fig:impact_tau} bottom. Therefore, a wider range of neighbors is included in the loss, and the model can significantly decrease the loss by maximizing distances to many relatively `easy negatives', negatives that are easily distinguishable from the anchor, \eg cat vs. parrot. 
By changing the temperature, the contribution of different negatives to the loss also changes, which leads to different semantic information being learned in the representation space. 
With small $\tau$, each individual sample maximizes the distance to its closest neighbor that results in a uniform space where the model learns instance specific features and each sample can be clearly distinguised from all others. 
With large $\tau$, only `easy negatives' are repelled from each other and the model becomes agnostic to instance-specific features, which allows the creation of semantic clusters, such that a `cat' cluster will be repelled from a `parrot' cluster. 

We follow \cite{kukleva2023temperature} and refer to learning with large $\tau$ as group-wise discrimination and learning with small $\tau$ as instance discrimination. 
Notably, allowing different types of discrimination is especially helpful in the case of long-tail distributions. 
In particular, tail classes need instance discrimination and benefit from local features, whereas head classes need group-wise discrimination and benefit from learned clusters. 


\begin{figure}
  \centering
  \includegraphics[width=1\linewidth]{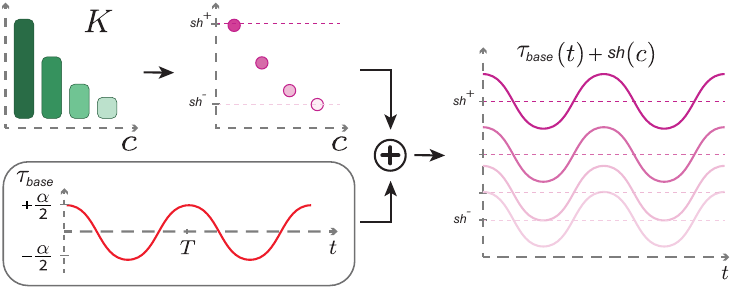}
  \caption{Detailed visualization of temperature calculation for every cluster $c$ on every training iteration $t$ based on the cluster distribution. Given the cosine schedule for the base temperature $\tau_{base}$ amplitude $\alpha$, oscillation period $T$ and cluster-based shifts $sh(c)$, we calculate temperature for each cluster.}
  \label{fig:schedule}
\end{figure}

\begin{figure*}[!t]
  \centering
  \begin{minipage}[t]{0.4\textwidth}
    \includegraphics[width=\textwidth]{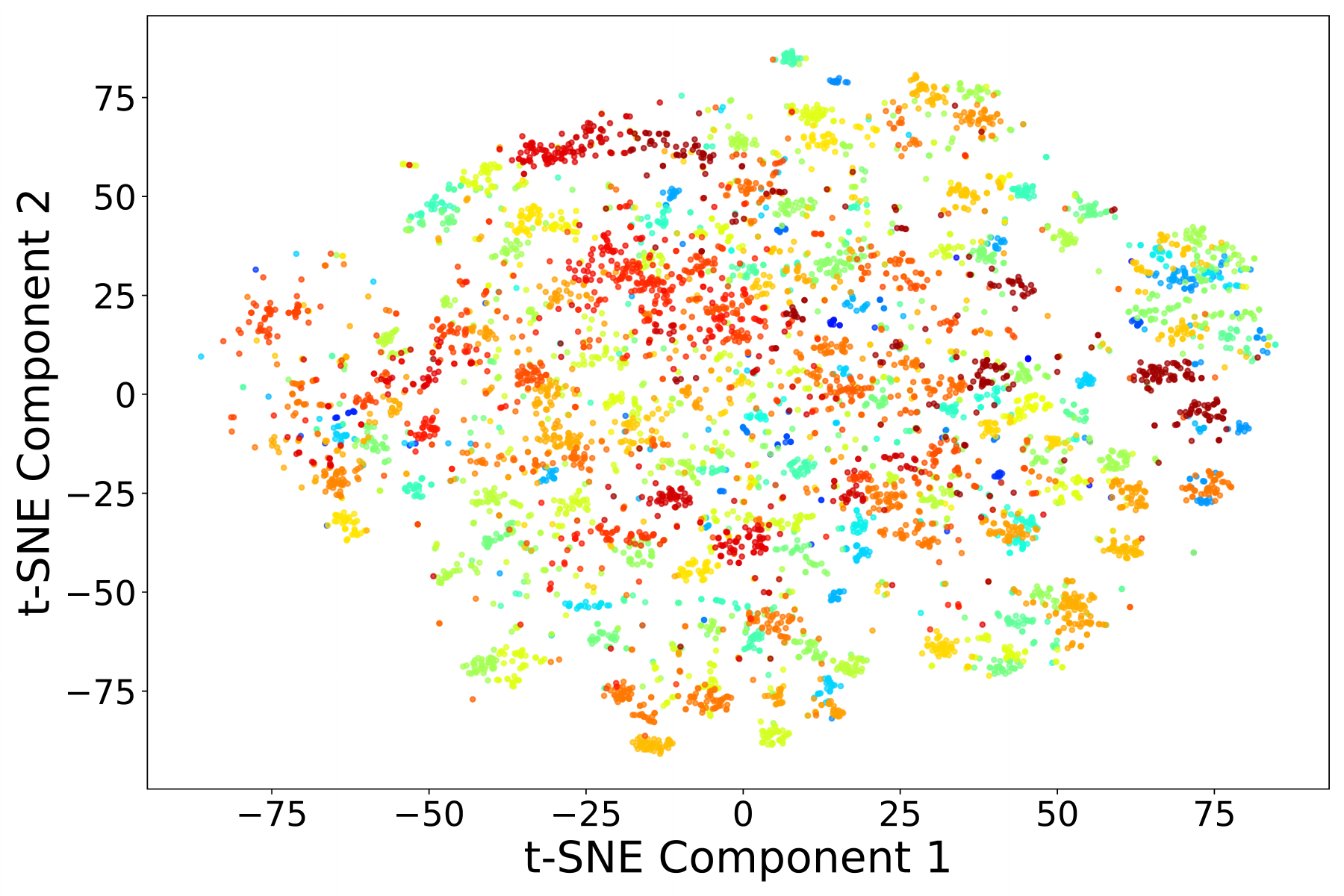}
    \caption{Visualization of the annotation embeddings in the YouCook2 dataset using tSNE. Each point represents a video annotation, and colors indicate the assigned clusters. Number of clusters is 200.}
    \label{fig:youcook_embeddings2}
  \end{minipage}
  \hfill
  \begin{minipage}[t]{0.51\textwidth}
    \includegraphics[width=\textwidth]{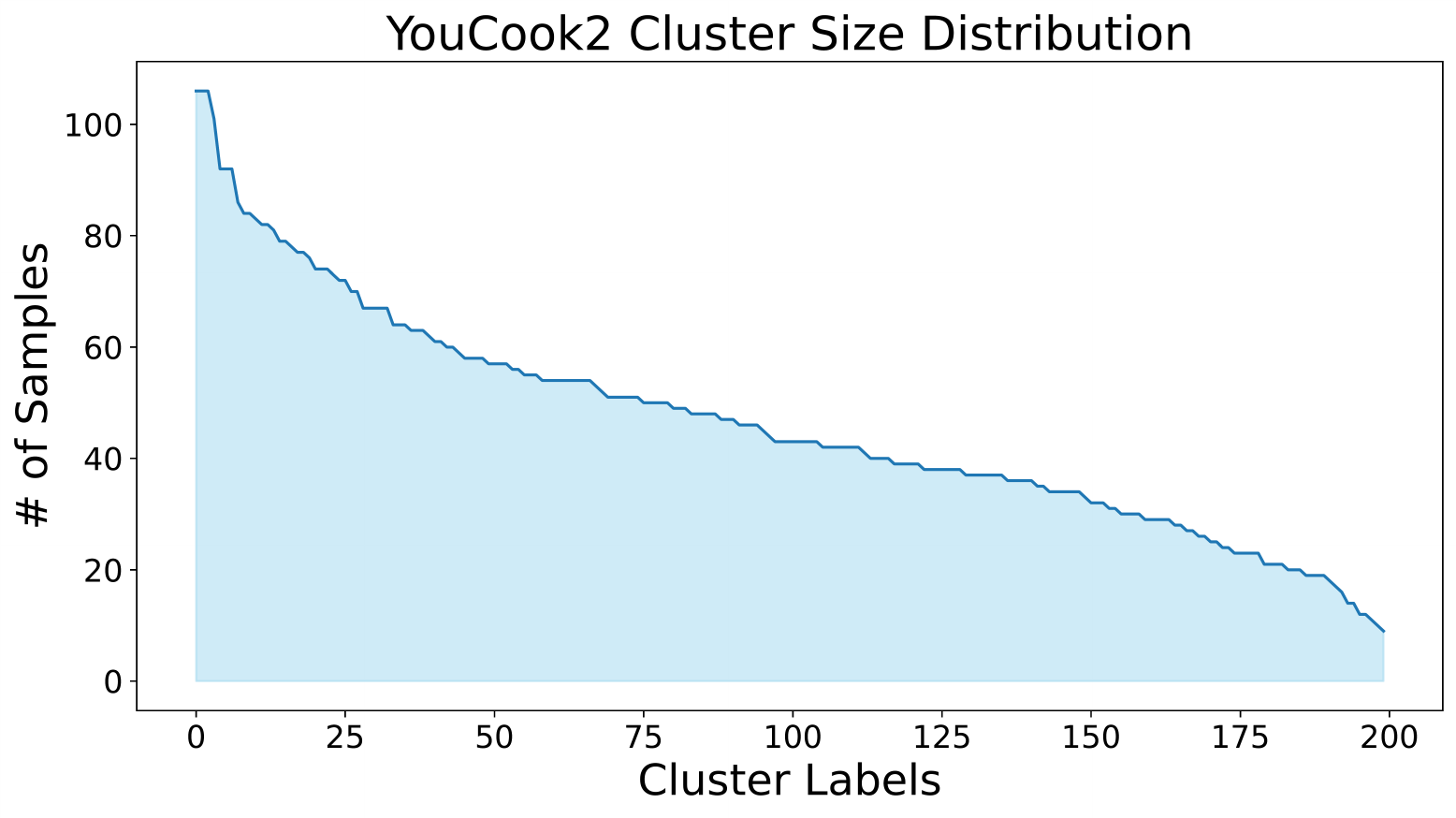}
    \caption{Visualization of long-tail annotations distribution of YouCook2 dataset. Annotations distribution is calculated based on k-mean clustering (200 clusters) of the annotation embeddings. Annotation embeddings are generated using SentenceBERT model \cite{reimers-2019-sentence-bert}.}
    \label{fig:youcook_dist}
  \end{minipage}
\end{figure*}

\myparagraph{Multi-Modal CL} includes at least two modalities, \eg images and their captions, where the objective is to increase similarity between the corresponding image-text pairs while repelling from all other instances of both modalities.   
In particular, there are two aligned sets of inputs, visual inputs $\{v_1,...,v_N\}$, \eg images or videos, and the corresponding textual descriptions $\{t_1,...,t_N\}$. 
Standard multi-modal contrastive learning includes two terms, from text to visual loss and from visual to text loss. 
Using visual $f_v$ and text $f_t$ encoders 
the multi-modal similarity scores are:
\begin{equation}
    s_{v\rightarrow t} = f_v(v)f_t(t)^T, \quad s_{t\rightarrow v} = f_t(t)f_v(v)^T.
    \label{eq:mm_cl_short}
\end{equation}
%
Multi-modal contrastive loss then is an average of the two InfoNCE losses (from vision to text and from text to vision):
\begin{equation}
    \frac{1}{2}  \left( \mathcal{L}_{\text{InfoNCE}}(s_{v\rightarrow t}) +  \mathcal{L}_{\text{InfoNCE}}(s_{t\rightarrow v}) \right).
    \label{eq:clip_loss}
\end{equation}
%
The key distinction between uni-modal and multi-modal contrastive learning lies in how samples interact within the embedding space. In uni-modal contrastive learning, each sample is compared against all other samples within the same modality. In contrast, multi-modal contrastive learning typically involves aligning samples across different modalities, where each instance is contrasted against all available samples from the other modality.
Despite these fundamental differences, we demonstrate that insights from uni-modal contrastive learning can be effectively leveraged to enhance multi-modal contrastive learning. 
Building on this, we propose a novel multi-modal contrastive learning framework that integrates key findings from uni-modal contrastive analysis.

\subsection{Multi-Modal Temperature Schedules}

Following our observations of the influence of $\tau$ on different modalities and findings in~\cite{kukleva2023temperature}, we introduce \underline{M}utli-\underline{M}odal \underline{T}emperature \underline{S}chedules, \acro, for long-tail multi-modal data distributions. 
\acro combines two components:
the first is the unsupervised temperature modulation, where we change temperature $\tau$ according to a cosine schedule similar to~\cite{kukleva2023temperature}, and the second is the individual temperature regulation.  
To estimate the 
ranges of $\tau$ for individual data points, we propose to leverage the aligned text and vision modalities in order to estimate the distribution of the training data. In particular, we propose to leverage pretrained language models to  approximate the distribution of the visual data using the corresponding text information. We show the pipeline to estimate the  temperature in \cref{fig:pipeline_main}.
 We then 
 use this approximation to estimate individual temperatures to enhance representation learning via a contrastive objective, ultimately improving model performance.

\begin{figure*}[!t]
\centering
\begin{subfigure}[b]{0.32\textwidth}
    \centering
    \includegraphics[width=0.82\textwidth]{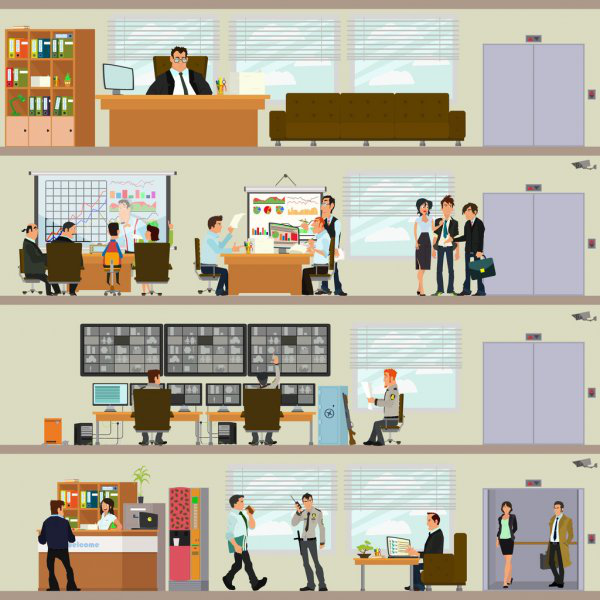}
    \caption{Scenes of \colorbox{yellow!40!white}{ people} working in the office.}
    \label{fig:cluster1}
\end{subfigure}
\hfill
\begin{subfigure}[b]{0.32\textwidth}
    \centering
    \vspace{-10mm}  
    \includegraphics[width=\textwidth]{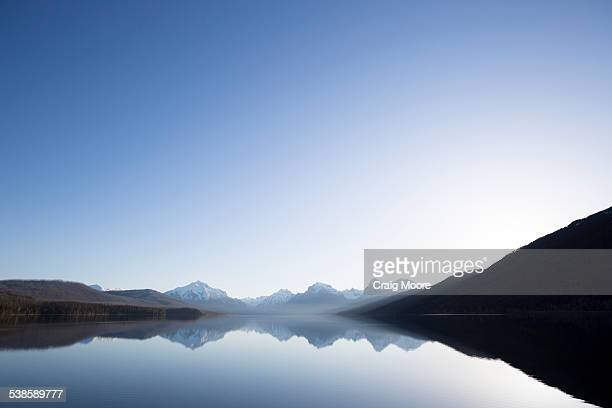}
    \vspace{0mm}    
    \caption{A calm morning before \colorbox{yellow!40!white}{sunrise}.}
    \label{fig:cluster2}
\end{subfigure}
\hfill
\begin{subfigure}[b]{0.32\textwidth}
    \centering
    \includegraphics[width=0.82\textwidth]{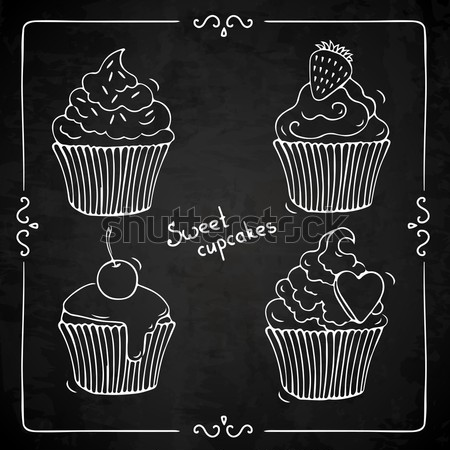}
    \caption{Sketches of cupcakes drawn on a \colorbox{yellow!40!white}{blackboard}.}
    \label{fig:cluster3}
\end{subfigure}
\caption{Qualitative samples from different estimated clusters for CC3M. The clusters are estimated using K-Means clustering. The corresponding cluster sizes from left to right are: 69669, 13708, 2904. With \colorbox{yellow!40!white}{yellow} we highlight manually identified topic for the corresponding cluster. For more text cluster examples see supmat. 
}
\label{fig:qualitative_CC3M}
\end{figure*}

\myparagraph{Individual temperature regulation.}
In \cite{kukleva2023temperature}, the authors discuss a toy experiment on long-tail unimodal data: overrepresented classes are assigned a large $\tau$ to encourage group-wise discrimination and learning of the larger clusters, whereas for underrepresented classes a small $\tau$ is assigned. The small $\tau$ results in instance discrimination and enforces underrepresented samples to push away from similar overrepresented classes improving the learned representations especially for these underrepresented classes.  
However, for unimodal settings with imbalanced data estimating the data distribution is a challenging. To overcome this problem, we leverage the multi-modal nature of the setting, where the text and visual data are pre-aligned, and  
extend the toy example from ~\cite{kukleva2023temperature} to the real data. 

\normalsize
Since the model is trained with contrastive learning to align text and visual modalities, the distribution of one modality can be used to approximate that of the other. In this work, we choose to approximate the distribution using text data, as text embeddings offer more robust semantic representations and are less affected by visual noise. Results of experiments using the visual modality as the distribution source are provided in the Supplementary Materials.
We extract all $N$ text embeddings of the corresponding text descriptions with the Bert~\cite{reimers-2019-sentence-bert} model and cluster these embeddings into $K$ clusters using K-Means. The size of the resulting clusters is used as a quantized frequency estimate for each corresponding data point. Larger sizes of the text clusters correspond to  larger values of $\tau$ used during training to induce larger semantic clusters in the multi-modal 
representation space. Whereas smaller text clusters indicate that some data is semantically underrepresented and could benefit from the explicit separation from other data points, therefore we assign a smaller $\tau$. 
More specifically, for a cluster $c$ with $K_c$ samples in the corresponding cluster,
the individual cluster shift of the respective temperature is calculated as: 
%
\small{
\begin{equation}
    sh(c) = \left(\frac{K_c- \text{min}(K)}{\text{max}(K) - \text{min}(K)} \right) \cdot (sh^{+} - sh^{-}) + sh^{-},
    \label{eq:tau-calc-prop}
\end{equation}}
%
%
%
%
%
\normalsize
where $sh^{+}$ and $sh^{-}$ define shifts for the largest and smallest clusters, respectively. 
Clustering and per-sample temperature adjustments introduce negligible overhead, as clustering occurs only once before training.
Inspired by temperature schedules~\cite{kukleva2023temperature}, we employ a similar intuition and define a dynamically changing schedule for the temperature with respect to the current iteration $t$ as follows:
\begin{equation}
    \tau_{base}(t) = \frac{\alpha \cos(2\pi t/T)}{2},
    \label{eq:tau-calc-corr}
\end{equation}
where $\alpha$ defines the range of changes of the base temperature. 
And the final temperature for every sample $i$ in the batch is calculated as a sum of the base temperature and the corresponding cluster shift $sh(c_i)$ for the current training iteration: 
%
\begin{equation}
    \tau_i = \tau_{base}(t) + sh(c_i),
    \label{eq:tau-calc-total}
\end{equation}
where $c_i$ is cluster 
of the corresponding sample $i$, see visualization in \cref{fig:schedule}.
Finally, we employ $\tau_i$ in the multi-modal contrastive loss \cref{eq:mm_cl_short}.



\myparagraph{Max-Margin Loss.}
We propose to extend our \acro idea to the max-margin loss.
In particular, we modify the margin $m$ in \cref{eq:max-margin} and substitute it with the modulated temperature defined as in~\cref{eq:tau-calc-total}. Therefore, we adjust margin in the loss for each sample individually and change it dynamicly over the training. 
We note that both contrastive losses operate using similar intuition behind them: pulling together similar samples while repelling dissimilar ones. 
Changes of the margin implies that with a small margin only the strongest negatives are pushed away, whereas with large margin those ``hard'' negatives do not impact the loss. 
However, in the max-margin loss the margin does not have influence on the `strength' of the positive relations in contrast to InfoNCE. Therefore, changing the margin affects only which negatives are pushed away. 
As we observe similar behavior of both losses, that further confirms 
that by controlling hardness of the negative samples it is possible to change the semantic of the embedding space.

\section{Experiments}
\label{sec:experiments}





In this section, we begin with a discussion of the datasets, the evaluation and the implementation details for the \acro method. Then, we discuss the distribution of each training dataset. Lastly, we show that  \acro improves the performance of various contrastive methods and ablate its components.

\subsection{Datasets}


\myparagraph{CC3M} (Conceptual Caption 3M)\cite{sharma2018conceptual} is a large-scale, internet-crawled image-text dataset. We use it for pretraining.
The training split consists of approximately 3.3M image-text pairs representing various styles.

\myparagraph{MS COCO} (Microsoft Common Objects in Context) is a large-scale dataset used for various tasks, including object detection and image captioning. It consists of 2.5M labels across 328K images, spanning 91 classes.
Inspired by previous works \cite{yuan2022provable, goel2022cyclip}, we use the Karpathy split \cite{karpathy2015deep} for our evaluation. It consists of 5K images for validation and another 5K for test. Retrieval at ranks 1, 5, and 10 is used for evaluation \cite{lin2014microsoft}.

\myparagraph{Flickr30k} includes 158K captions across 30K images in different environments. Similar to MS COCO, we use the Karpathy split with 1K images for validation and another 1K for testing. Retrieval at ranks 1, 5, and 10 is used for evaluation \cite{plummer2015flickr30k}.

\myparagraph{EPIC-KITCHENS-100} (EK-100) is a large-scale video dataset, containing egocentric videos with kitchen-based activities. In our work, we focus on the Multi-Instance Retrieval (MIR) task, which is a video-text retrieval extended with sematic relevancy between narrations. The dataset contains 97 verb and 300 noun classes. The train split contains 67.2K annotated segments, validation split -  9.6K.
\cite{damen2022rescaling},

\myparagraph{YouCook2} is a large-scale video dataset focusing on cooking activities. The dataset contains mainly the 3rd person view videos. Training split contains 10K annotated video segments, validation split contains  3.5K annotated segments. \cite{zhou2018towards}




\begin{table}[t!]
    \centering
    \resizebox{\linewidth}{!}{%
    \setlength{\tabcolsep}{1.8pt}
    \begin{tabular}{l|l|c|c|c|c|c|c}
        \multicolumn{1}{c|}{\multirow{2}{*}{Method}} 
        &  Back- & \multicolumn{3}{c|}{Flickr30K} & \multicolumn{3}{c}{MSCOCO} \\
        &  bone  & IR@1 & TR@1 & Avg. & IR@1 & TR@1  & Avg.  \\ \Xhline{3\arrayrulewidth}
        CLIP~\cite{radford2021learning} & RN50 & 40.9 & 50.9 & 45.9 & \textbf{21.3} & 26.9 & 24.1 \\
        \cellcolor[HTML]{EFEFEF}{CLIP + Ours} & \cellcolor[HTML]{EFEFEF}{RN50} & \cellcolor[HTML]{EFEFEF}{\textbf{41.5}} & \cellcolor[HTML]{EFEFEF}{\textbf{54.3}} & \cellcolor[HTML]{EFEFEF}{\textbf{47.9}} & \cellcolor[HTML]{EFEFEF}{21.2} & \cellcolor[HTML]{EFEFEF}{\textbf{28.4}} & \cellcolor[HTML]{EFEFEF}{\textbf{24.8}} \\ 
        \hline
    \end{tabular}
    }
    \caption{Performance comparison of pretraining on CC3M and zero-shot evaluation on MSCOCO and Flickr30K.  We combine CLIP with our method. }
    \label{tab:CC3M}
    \vspace{-3mm}
\end{table}

\subsection{Evaluation}
Our work addresses two closely related tasks: Image/Video Retrieval and MIR. Multi-Instance Retrieval  further necessitates a relevancy matrix to capture the relationships between samples in the dataset. Consequently, we apply two distinct sets of evaluation metrics: Mean Average Precision (mAP) and normalized Discounted Cumulative Gain (nDCG) for MIR, and Recall@1,5,10 for Text-Image/Video Retrieval. Evaluations are conducted bidirectionally, assessing both text-to-image/video and image/video-to-text retrieval performance. For EK-100 and YouCook2, we  fine-tune and evaluate on the respective datasets. Whereas, for CC3M experiments, we first fine-tune on CC3M and then perform zero-shot retrieval on MS COCO and Flick30K.

\subsection{Implementations Details}

For experiments on CC3M, we use standard CLIP~\cite{radford2021learning} framework and conduct experiments with ResNet50 architecture. We initialize it with ImageNet-1K pretrained model. For the CLIP baseline, we use fixed temperature 0.01 as it performed the best.



We conduct all experiments on the EK-100 dataset following the AVION~\cite{zhao2023training} protocol for fine-tuning on the EK-100 MIR task. 



For the YouCook2 dataset, we use the fine-tuning configuration of the VAST~\cite{chen2023vast} model for the retrieval task. For the calculation of the cluster distribution we first calculate embeddings of the text annotations using SentenceBert model \cite{reimers-2019-sentence-bert} and then find the clusters with K-Means ($k=200$). 


Regarding hyperparameters ($\alpha$, $sh^{-}$, $sh^{+}$), our experiments demonstrate optimal results when parameters oscillate around default temperature or margin values. Specifically, we chose $\alpha$ to keep temperatures positive ($\tau_{\text{default}} - \alpha/2 > 0$), and set $sh^{-}$ and $sh^{+}$ within $[\tau_{\text{default}}-\alpha/2, \tau_{\text{default}}+\alpha/2]$, preventing negative values for tail samples. The values can be found in Sec. \ref{sec:hyperparameters} (supmat).

\begin{table}[t!]
\centering
\resizebox{\linewidth}{!}{%
\setlength{\tabcolsep}{1.8pt}
\begin{tabular}{l|l|c|c|c|c|c|c}
\multicolumn{1}{c|}{\multirow{2}{*}{Method}} 
&  Back- & \multicolumn{3}{c|}{mAP} & \multicolumn{3}{c}{nDCG} \\
&  bone  & V$\rightarrow$T & T$\rightarrow$V & Avg. & V$\rightarrow$T & T$\rightarrow$V & Avg.  \\ \Xhline{3\arrayrulewidth}
MME \cite{wray2019fine} & TBN & 43.0 & 34.0 & 38.5 & 50.1 & 46.9 & 48.5 \\
JPoSE \cite{wray2019fine} & TBN & 49.9 & 38.1 & 44.1 & 55.5 & 51.6 & 53.5 \\
EgoVLP \cite{lin2022egocentric} & TSF-B & 49.9 & 40.5 & 45.0 & 60.9 & 57.9 & 59.4 \\
LaViLa \cite{zhao2023learning} & TSF-B & 55.2 & 45.7 & 50.5 & 66.5 & 63.4 & 65.0 \\
AVION \cite{zhao2023training} & ViT-B & 55.7 & 48.2 & 52.0 & 67.8 & 65.3 & 66.5 \\ 


\cellcolor[HTML]{EFEFEF}{AVION + Ours}& \cellcolor[HTML]{EFEFEF}{ViT-B} & \cellcolor[HTML]{EFEFEF}{\textbf{58.8}} & \cellcolor[HTML]{EFEFEF}{\textbf{48.9}} & \cellcolor[HTML]{EFEFEF}{\textbf{53.9}} & \cellcolor[HTML]{EFEFEF}{\textbf{68.9}} & \cellcolor[HTML]{EFEFEF}{\textbf{65.8}} & \cellcolor[HTML]{EFEFEF}{\textbf{67.3}} \\  \hline 
\end{tabular}
}
\caption{Performance comparison of MIR on EPIC-KITCHENS-100 dataset. We combine AVION with our method. 
}
\label{tab:EK_sota}
\end{table}

\subsection{Estimating the Distribution}
The annotations of the EK-100 dataset include extracted verbs and nouns, the dataset naturally follows a long-tail distribution. We visualize verb, noun, and verb-noun combination distributions in \cref{fig:ek100_distributions} and use these three variants as class distribution sources in our experiments.

For the YouCook2 and CC3M datasets, we derive the distribution of the training data from the K-Means cluster distribution of annotation embeddings. For this purpose, we utilize the Sentence-BERT \cite{reimers-2019-sentence-bert} and distilled version of BERT \cite{sanh2019distilbert}, respectively. With K-Means we estimate 200 distinct clusters, we visualize the estimated distributions in \cref{fig:youcook_dist,fig:cc3m_class_distribution}. 
In \cref{fig:qualitative_CC3M}, we show random qualitative samples from clusters of different sizes from CC3M dataset. 
As these datasets are collected without following any specified class-distribution, we observe that the formed clusters are indeed follow long-tail distribution. In \cref{fig:youcook_embeddings2,fig:cc3m_embeddings2}, we plot tSNE visualization of the clusters for the YouCook2 and CC3M datasets. We observe that most of the clusters has easily identifiable topic, such as \textit{hummus} or \textit{grill}, see \cref{tab:clusters_good_youcook,tab:clusters_topcc3m,tab:clusters_bottom_cc3m} (supmat). 



\subsection{Results}
We compare the \acro method with current state-of-the-art methods. First, in \cref{tab:CC3M}, we compare our \acro method with a standard CLIP objective. We observe that our dynamic individualized temperatures are especially helpful for text-to-image retrieval improving by 3.4\% on zero-shot evaluation on Flickr30K and 1.5\% on MSCOCO. 
Second, in \cref{tab:EK_sota}, we evaluate the performance on EPIC-KITCHENS-100 dataset using max-margin loss, as it was shown~\cite{zhao2023learning, lin2022egocentric} that max-margin loss performs significantly better than InfoNCE when the model is fine-tuned on EK-100. We extend AVION framework based on max-margin loss with the \acro framework for dynamic margin estimation and observe improvments across all metrics achieving new state-of-the-art on EK-100. Notably, mAP $V\rightarrow T$ improves more than by 3\%.  
Third, in \cref{tab:youcook2_sota}, we present a comparison of text-to-video retrieval performance on the YouCook2 dataset based on the VAST~\cite{chen2023vast} method. Our method achieves significant performance improvements, surpassing the original VAST results by 2.2-4\% and achieving new sota on YouCook2 dataset retrieval task as well.

\begin{table}[t!]
\centering
\begin{tabular}{l|c|c|c}
\multicolumn{1}{c|}{\multirow{2}{*}{Method}}  & \multicolumn{3}{c}{Text-to-Video} \\
                & R@1 & R@5 & R@10 \\ \hline
UniVL \cite{luo2020univl}  & 28.9 & 57.6 & 70.0 \\
MELTR \cite{ko2023meltr}  & 33.7 & 63.1 & 74.8 \\
VLM \cite{xu-etal-2021-vlm}   & 27.1 & 56.9 & 69.4 \\ 
VAST \cite{chen2023vast} & 50.4  & 74.3  & 80.8 \\
\cellcolor[HTML]{EFEFEF}{VAST + Ours} & \cellcolor[HTML]{EFEFEF}{\textbf{53.0}}   & \cellcolor[HTML]{EFEFEF}{\textbf{77.1}} & \cellcolor[HTML]{EFEFEF}{\textbf{84.5}} \\ \hline

\end{tabular}
\caption{Performance comparison on the YouCook2 dataset. We combine VAST with our method.  The evaluation is done with Text-to-Video Recall@1,5,10. All benchmarks are multi-modal, containing both audio and subtitles. Methods that also incorporate audio and subtitle modalities (in addition to vision for video representation) are highlighted with a gray background.}
\label{tab:youcook2_sota}
\vspace{-3mm}
\end{table}




\subsection{Ablations}

\myparagraph{Impact of \acro.}
In \cref{tab:EK_avion_ablations}, we analyze the impact of the two main components of our method: temperature (margin) schedules (TS) and individual cluster shifts (ICS), on EK-100 based on MIR task. 
In particular, TS is the dynamic change of the temperature that follows cosine schedule, whereas ICS are shifts that we employ for the base temperature to adjust for each individual sample. 
In \cref{tab:EK_avion_ablations}, we evaluate both, the modified max-margin (MI-MM) and the InfoNce loss (CLIP). We observe that TS is especially effective on the CLIP loss as it improves by 4.3\% and 4.0\% on avg. mAP and avg. nDCG, correspondingly, whereas ICS is beneficial for MI-MM (max-margin) loss improving by 0.9\% on avg. nDCG. However, the combination of the two components brings the best improvements on average for both losses. 

\begin{table}
\centering
\setlength{\tabcolsep}{4pt}
\begin{tabular}{l|c|c|c|c|c|c}
\multicolumn{1}{c|}{\multirow{2}{*}{Method}} 
 & \multicolumn{3}{c|}{mAP} & \multicolumn{3}{c}{nDCG} \\
& V$\rightarrow$T & T$\rightarrow$V & Avg. & V$\rightarrow$T & T$\rightarrow$V & Avg.  \\ \Xhline{3\arrayrulewidth}
MI-MM  & 55.3 & 47.6 & 51.4 & 67.6 & 64.8 & 66.2 \\ 
w/ TS  & 58.4 & 48.6 & 53.6 & \textbf{69.4} & 64.8 & 66.9 \\
w/ ICS &  56.4 & 48.4 & 52.4 & 68.2 & \textbf{66.0} & 67.1 \\
w/ TS\&ICS &  \textbf{58.8} & \textbf{48.9} & \textbf{53.9} & 68.9 & 65.8 & \textbf{67.3} \\ \hline
CLIP          & 47.0 & 39.8 & 43.1 & 64.4 & 61.2 & 62.8 \\ 
w/ TS     & \textbf{55.0} & 41.7 & 48.4 & \textbf{69.1} & 64.6 & \textbf{66.8} \\
w/ ICS     &  52.2  &  43.6  & 47.8   &  68.0  &  \textbf{64.8}  &  66.4    \\
w/ TS\&ICS & 54.6 & \textbf{43.9} & \textbf{49.2} & 68.6 & 64.4 & 66.4 \\
\hline

\end{tabular}
\caption{ Performance comparison of AVION on EK-100 dataset (MIR) with modified max-margin- (MI-MM) and CLIP- (InfoNCE) losses. TS and ICS temperature (or margin) schedule and individual cluster shifts, respectively. }
\label{tab:EK_avion_ablations}
\end{table}



\myparagraph{Distribution estimation. }
For EK-100 class annotations such as nouns, verbs and actions are already available (see the corresponding distributions in \cref{fig:ek100_distributions} (supmat)), with an action being a combination of a respective noun and verb. In \cref{tab:EK_avion_verbs_nouns_both}, we evaluate influence of the different distributions on the individual cluster shift estimation. We observe that distribution estimated based on nouns results in the best performing model improving both mAP and nDCG results. Interestingly, we manually check formed clusters for both CC3M and YouCook2 (see in Tabs. \ref{tab:clusters_good_youcook}, \ref{tab:clusters_topcc3m}, \ref{tab:clusters_bottom_cc3m} supmat) and notice that the clusters' topics are biased towards objects and do not reflect distribution of actions (for YouCook2). 

\begin{table}
\centering
\setlength{\tabcolsep}{3pt}
\begin{tabular}{l|c|c|c|c|c|c}
\multicolumn{1}{c|}{\multirow{2}{*}{\makecell{Distribution \\ source}}}  & \multicolumn{3}{c|}{mAP} & \multicolumn{3}{c}{nDCG} \\
& V$\rightarrow$T & T$\rightarrow$V & Avg. & V$\rightarrow$T & T$\rightarrow$V & Avg.  \\ \Xhline{3\arrayrulewidth}
verbs \& nouns &  58.2 & 48.7 & 53.4 & 68.3 & \textbf{65.5} & 66.9 \\
verbs  & 58.5 & 48.7 & 53.6 & 68.0 & 65.3 & 66.7 \\
nouns &  \textbf{58.8} & \textbf{48.9} & \textbf{53.9} & \textbf{68.9} & 65.3 & \textbf{67.3} \\ 
\hline

\end{tabular}
\caption{ Performance comparison of AVION with margin distribution method based on the distribution of uniq verbs+nouns combinations, verbs and nouns}
\label{tab:EK_avion_verbs_nouns_both}
\vspace{-3mm}
\end{table}

\section{Conclusion}

In this work, we proposed Multi-Modal Temperature and Margin Schedules (\acro), a framework designed to enhance multi-modal contrastive learning on long-tail data. Inspired by uni-modal temperature schedules \cite{kukleva2023temperature}, we extended this concept to the multi-modal setting.
First, we introduced a dynamic temperature schedule that follows a cosine schedule for both modalities. This approach enables the model to capture different types of semantic information: higher temperatures facilitate the formation of semantic groups in the embedding space, while lower temperatures enhance instance discrimination, helping to distinguish similar samples based on local semantics.
To further amplify this effect, we proposed incorporating distribution-dependent shift values. Specifically, we estimated the data distribution either using available class labels or by applying K-Means clustering to pre-extracted text embeddings. Notably, we observed that the resulting clusters encode object-centric semantic structures for both image- and video-language datasets, improving the alignment between modalities.
We evaluated our approach on image- and video-language datasets using both InfoNCE and Max-Margin losses. Our results demonstrated consistent improvements across all datasets, evaluation metrics, and losses, highlighting the effectiveness of our method in multi-modal contrastive learning on long-tail data.

{
    \small
    \bibliographystyle{ieeenat_fullname}
    \bibliography{main}
}
\clearpage
\setcounter{page}{1}
\maketitlesupplementary

In the supplementary, we provide full formulation of the multi-modal contrastive InfoNCE loss,
additional ablations, results on additional dataset SSv2-LT, and insights of the clusters for CC3M and YouCook2 datasets. 

\section{Mutli-Modal InfoNCE Loss }
Given the multi-modal similarity scores:
\begin{equation}
    s^{v\rightarrow t} = f_v(v)f_t(t)^T, \quad s^{t\rightarrow v} = f_t(t)f_v(v)^T,
    \label{eq:mm_cl_short}
\end{equation}
%
The $\text{InfoNCE}(s_{t\rightarrow v})$ can be formulated as the following:
\begin{equation}
   \mathcal{L}_{\text{InfoNCE}(s_{t\rightarrow v})} = -\frac{1}{N} \sum_{i=1}^{N} \log \frac{\exp(s^{t\rightarrow v}_{ii} / \tau)}{ \sum_{j=1}^{N} \exp(s^{t\rightarrow v}_{ij} / \tau)}.
  \label{eq:info-nce}
\end{equation}
And the $\text{InfoNCE}(s_{v\rightarrow t})$ can be formulated as the following:
\begin{equation}
   \mathcal{L}_{\text{InfoNCE}(s_{v\rightarrow t})} = -\frac{1}{N} \sum_{i=1}^{N} \log \frac{\exp(s^{t\rightarrow v}_{ii} / \tau)}{ \sum_{j=1}^{N} \exp(s^{v\rightarrow t}_{ij} / \tau)}.
  \label{eq:info-nce}
\end{equation}

The the multi-modal contrastive loss is an average of the two InfoNCE losses (from vision to text and from text to vision):
\begin{equation}
    \frac{1}{2}  \left( \mathcal{L}_{\text{InfoNCE}}(s_{v\rightarrow t}) +  \mathcal{L}_{\text{InfoNCE}}(s_{t\rightarrow v}) \right)
    \label{eq:clip_loss}
\end{equation}

\section{Hyperparameters}
\label{sec:hyperparameters}

We provide the exact values of the hyperparameters $\alpha$, $sh^{-}$, and $sh^{+}$ used in the experiments. These values were selected according to the criteria outlined in the main text, ensuring positive temperatures and margins while maintaining the intended oscillation around the default values.

\begin{itemize}
    \item \textbf{\acro + CLIP (CC3M):} \\ $\alpha = 0.04$, $sh^- = 0.05$, $sh^+ = 0.10$
    \item \textbf{AVION (MI-MM loss):} \\ $\alpha = 0.20$, $sh^- = 0.17$, $sh^+ = 0.30$
    \item \textbf{AVION (CLIP loss):} \\ $\alpha = 0.08$, $sh^- = 0.05$, $sh^+ = 0.20$
    \item \textbf{VAST (MI-MM loss):} \\ $\alpha = 0.20$, $sh^- = 0.10$, $sh^+ = 0.30$
    \item \textbf{VAST (CLIP loss):} \\ $\alpha = 0.06$, $sh^- = 0.06$, $sh^+ = 0.09$
\end{itemize}


\section{Component Analysis of \acro on YC2}

Additionally, we provide breakdown influence of the components of our \acro method on YouCook2 dataset. Note, that VAST\cite{chen2023vast} has two types of evaluation with and without refinement. Without refinement evaluation is similar to the standard retrieval evaluation as with CLIP and AVION, whereas refinement includes additional training modules. In \cref{tab:YC2_vast_ablations}, we evaluate performance of VAST with InfoNCE (CLIP) and Max-Margin (MI-MM) losses. We present results with and without refinement. 

\section{Modality Choice for Distribution Estimation}

We provide comparison training based on clusters from text embeddings (Sentence-BERT) and video embeddings (VAST backbone) on YouCook2 (see S-BERT vs VAST-v in \cref{tab:distribution_source}). Notably, we find that video clusters also exhibit a similar long-tail distribution and modulation based on video clusters shows comparable improvement over the VAST baseline. Therefore, we conclude that while any modality can be effectively used for modulation, text is generally preferable when available. 
\begin{table}[h]
\centering
    \begin{tabular}{c|ccc}
         & R@1 & R@5 & R@10\\ \hline
         VAST (baseline) & 50.4 & 74.3 & 80.8\\
        S-BERT(text) & \textbf{53.0} & \textbf{77.1} & \textbf{84.5}\\
        VAST-v(video)  & 52.2 & 76.3 & 84.0 \\
        CLIP-t(text) & 52.5 & 76.4 & 84.4 \\
        \bottomrule
    \end{tabular}
    \caption{Comparison of retrieval performance on YouCook2 using distribution modulation based on text (S-BERT, CLIP-t) and video (VAST-v) clusters.}
    \label{tab:distribution_source}
\end{table} 

\section{Something-Something-v2-LT dataset}

Additionally, we extend our method to complex Something-Something-v2 (SSv2)~\cite{goyal2017something} dataset. The dataset includes 169K training videos and 25K validation videos, each depicting interactions with everyday objects. The dataset has 174 action classes. 
SSv2-LT is a long-tailed subset of the original Something-Something-v2 dataset, proposed by Perrett \etal \cite{perrett2023use}. The resampling of training split was done using Pareto distribution with $\alpha=6$. Validation and test splits are balanced and contain 40 and 15 samples per class respectively~\cite{perrett2023use}. In~\cite{perrett2023use}, the authors use the standard classes for the classification task, whereas we rely on the given distribution to estimate individual shift values. Then, we train the model using full captions (without ``something'' placeholder) and evaluate on text and video retrieval tasks. As a base framework, we use VAST. In \cref{tab:vast_SSv2LT_results}, we additionally evaluate all the components of our \acro framework and show the improvements with the combination of the temperature schedules and individual adjustments. 

Moreover, in \cref{tab:ablation_temp_amp}, we evaluate how robust our method to different temperatures. Our results demonstrate that the method is robust to variations of the temperature. 

\begin{table}[h]
\centering
\setlength{\tabcolsep}{3.5pt}
\begin{tabular}{l|c|c|c|c|c|c}
\multicolumn{1}{c|}{\multirow{2}{*}{Method}} 
& \multicolumn{3}{c|}{Text-to-Video} & \multicolumn{3}{c}{Video-to-Text} \\ 
            & R@1 & R@5 & R@10 & R@1 & R@5 & R@10 \\ \Xhline{3\arrayrulewidth}

VAST \cite{chen2023vast} &  41.7  & 69.1  & 78.9  & 41.2 &  69.3  &  79.3  \\ 
w/ TS   &  42.6  &  70.0  &  79.5  &  41.3  &  68.7  &  79.3   \\ 
w/ TD   & 42.7 &  \textbf{70.3} &  79.6  &  41.2  &  69.7  &  79.5  \\ 
w/ TS\&TD & \textbf{43.2}  &  \textbf{70.3}  & \textbf{79.7}  & \textbf{41.8}  &  \textbf{70.3}  & \textbf{80.0}  \\ \hline
\end{tabular}
\caption{Performance comparison of variants of VAST on finetuning on SSv2-LT  dataset. Results without refinement.}
\label{tab:vast_SSv2LT_results}
\end{table}

\begin{table}[h]
\centering
\setlength{\tabcolsep}{4.5pt}
\begin{tabular}{c|c| c| c| c}
\diagbox{$\tau_{\downarrow}/\tau_{\uparrow}$}{$\alpha$} & 0.02 & 0.04 & 0.06 & 0.08  \\ \hline
0.05/0.09  & 42.7 & 43.2  & 43.2 & 42.3 \\ 
0.06/0.10  & \textbf{43.4} & 43.0 & 43.1 & 42.8  \\ 
0.07/0.11  & 43.0 & 43.2 & 42.9  & 42.8   \\ \hline
\end{tabular}
\caption{Impact of different temperature ranges ($\tau_{min}$ represented with $\tau_{\downarrow}$ and $\tau_{max}$ represented with $\tau_{\uparrow}$) and amplitude ($\alpha$) values on the performance of VAST on SSv2-LT, evaluating R@1 Text-to-Video. The combinations marked with ``-'' were not done as the resulting lowest temperature for the smallest class would be $\leq$ 0.}
\label{tab:ablation_temp_amp}
\end{table}


\section{Text Distributions}

\begin{figure*}[t!]
  \centering
  \begin{minipage}[t]{0.4\textwidth}
    \includegraphics[width=\textwidth]{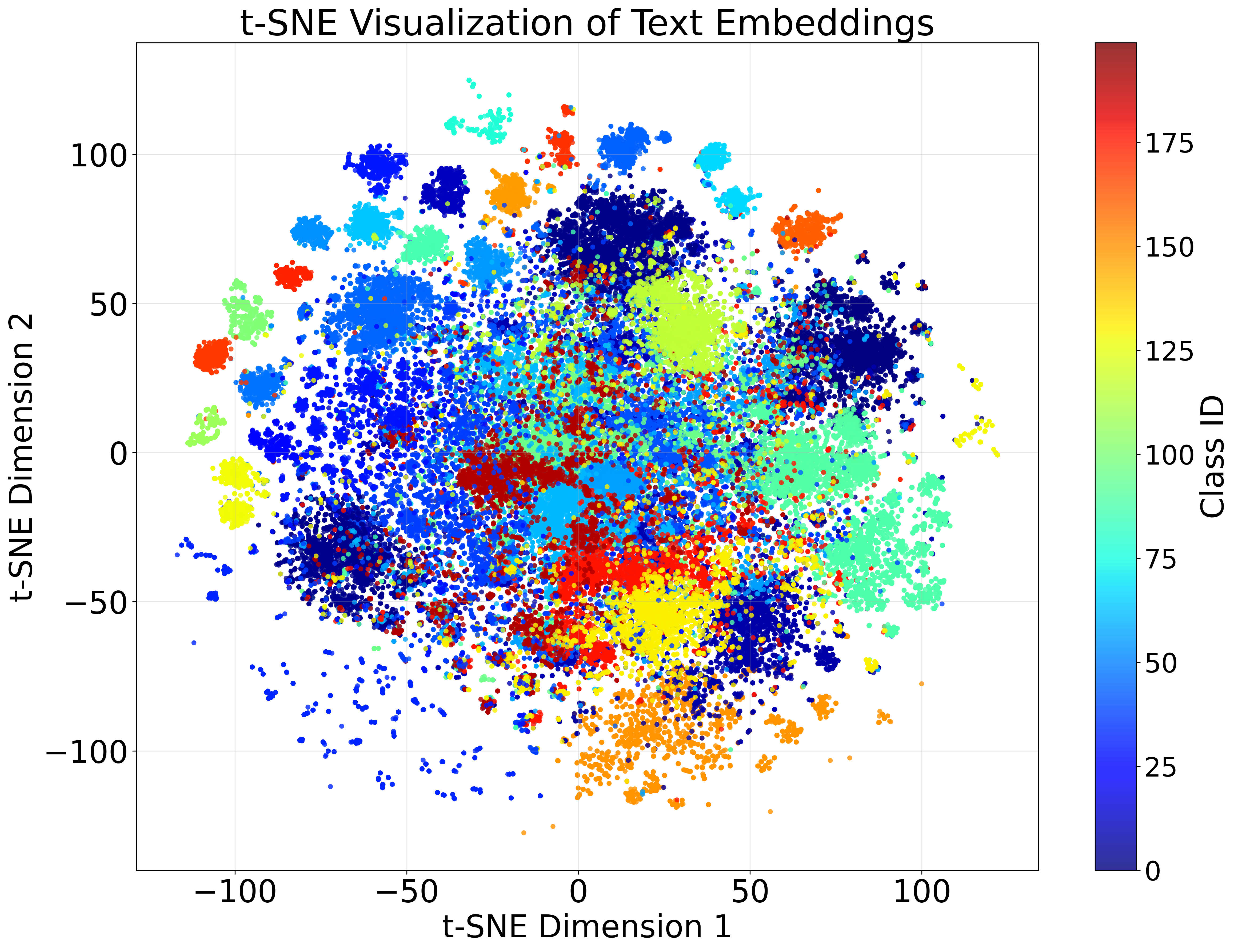}
    \caption{Visualization of the annotation embeddings in the CC3M dataset using tSNE. Each point represents a image annotation, and colors indicate the assigned clusters. In our experiments, we used 200 clusters.}
    \label{fig:cc3m_embeddings2}
  \end{minipage}
  \hfill
  \begin{minipage}[t]{0.55\textwidth}
    \includegraphics[width=\textwidth]{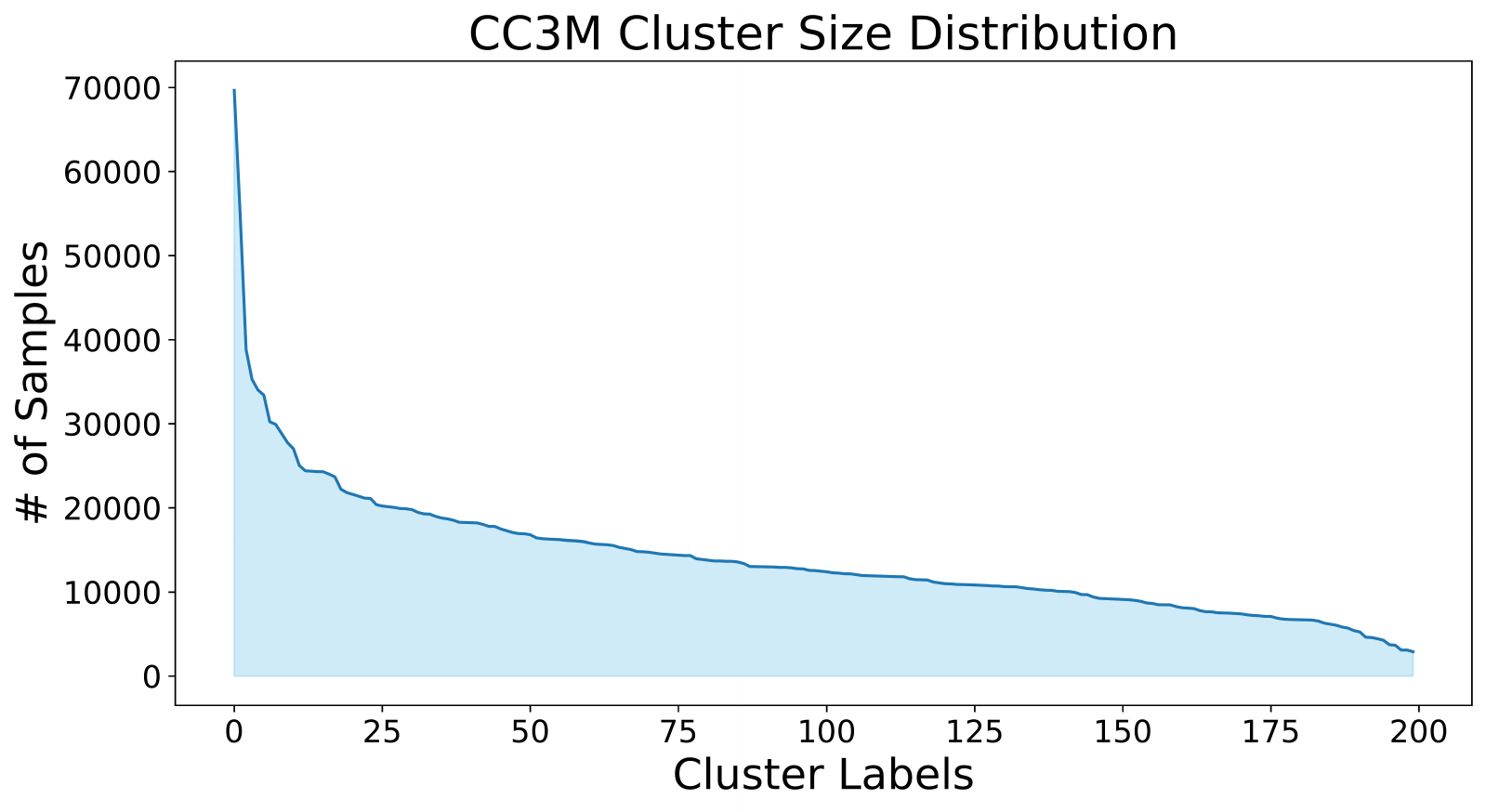}
    \caption{Visualization of long-tail annotations distribution of CC3M
dataset. Annotations distribution is calculated based on k-mean clustering(200 clusters) of the annotation embeddings. Annotation embeddings are generated using BERT model~\cite{sanh2019distilbert}.}
\label{fig:cc3m_class_distribution}
  \end{minipage}
\end{figure*}

\begin{figure*}
  \centering
  \includegraphics[width=1\linewidth]{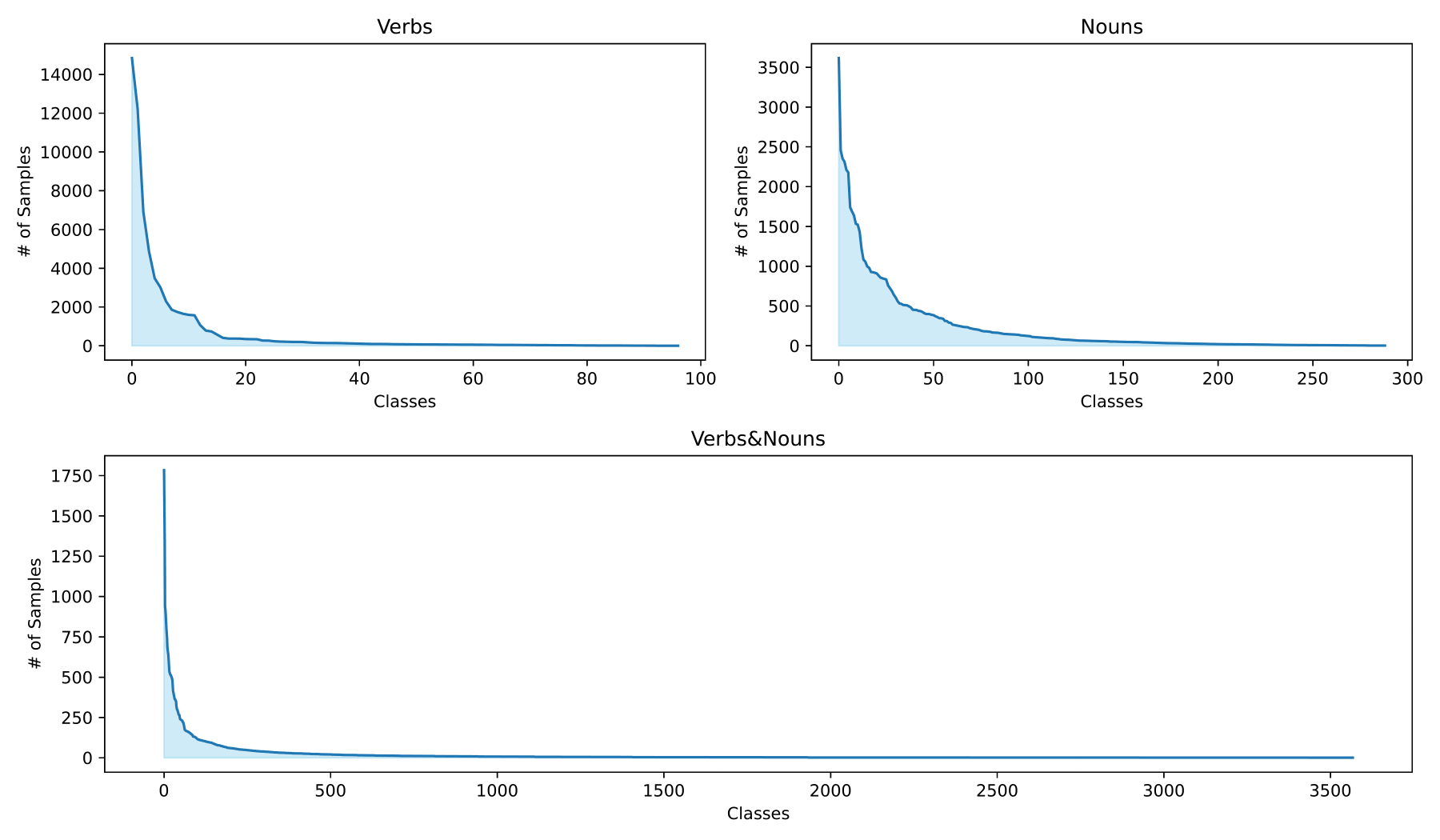}
  \caption{Visualizations of long-tail class distributions of Epic-Kitchens100 dataset training split. Class distributions are calculated based on verbs (top-left), important nouns (top-right) and unique combinations of verbs and nouns (bottom).}
  \label{fig:ek100_distributions}
\end{figure*}

\begin{figure}
  \centering
  \includegraphics[width=1\linewidth]{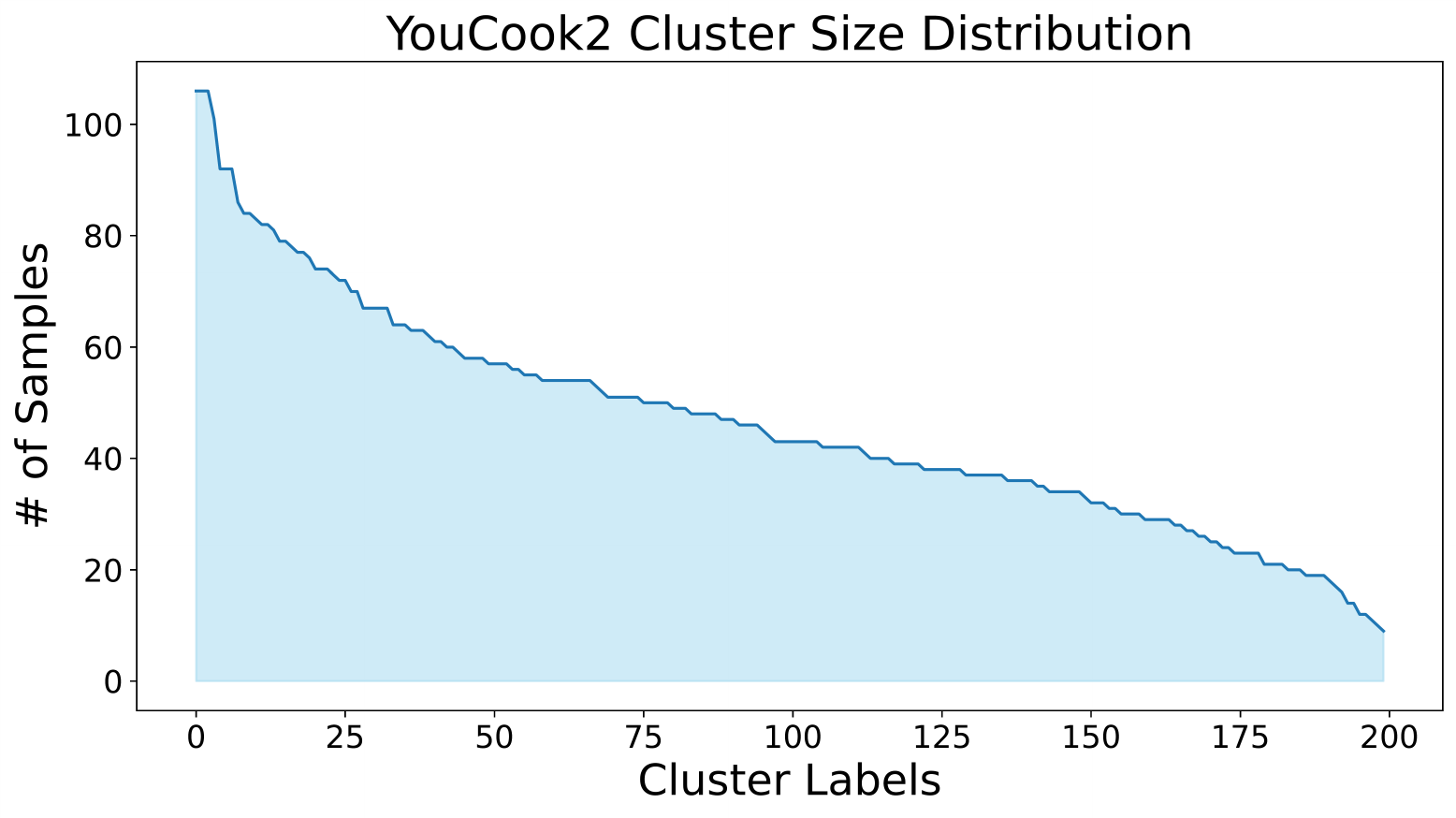}
  \caption{Visualization of long-tail annotations distribution of YouCook2 dataset. Annotations distribution is calculated based on k-mean clustering (200 clusters) of the annotation embeddings. Annotation embeddings are generated using SentenceBERT model \cite{reimers-2019-sentence-bert}.}
  \label{fig:youCook2_distributions}
\end{figure}

\begin{figure}
  \centering
  \includegraphics[width=1\linewidth]{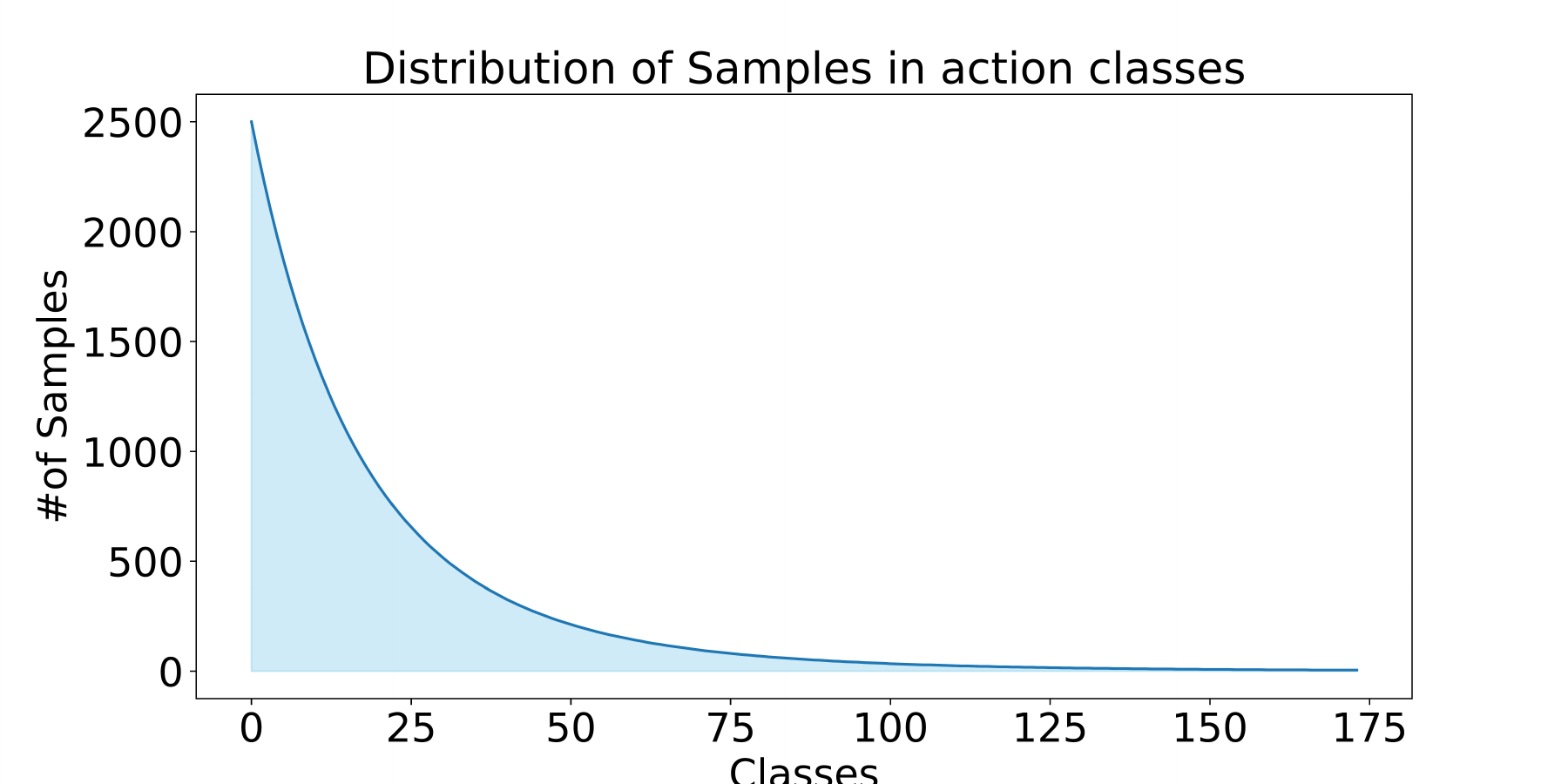}
  \caption{Visualization of action class distribution of SSv2-LT dataset. }
  \label{fig:ssv2_distributions}
\end{figure}

In Figs.~\ref{fig:cc3m_class_distribution}, \ref{fig:youCook2_distributions}, we plot distributions based on 200 estimated clusters in the text embeddings for CC3M and YouCook2 datasets, respectively. 
In Figs.~\ref{fig:ek100_distributions},  \ref{fig:ssv2_distributions}, we plot class-based distributions for EPIC-KITCHENS-100, and SSv2-LT datasets, respectively. In Tabs. \ref{tab:clusters_good_youcook}, \ref{tab:clusters_topcc3m}, \ref{tab:clusters_bottom_cc3m}, we show random examples of captions in different clusters and manually identify the common topic between the sampled sentences.

\begin{table}[h]
\centering
\setlength{\tabcolsep}{4pt}
\begin{tabular}{l|c|c|c|c|c|c}
\multicolumn{1}{c|}{\multirow{2}{*}{Method}} 
 & \multicolumn{3}{c|}{Text-to-Video} & \multicolumn{3}{c}{Video-to-Text} \\
& R@1 & R@5 & R@10 & R@1 & R@5 & R@10   \\ \Xhline{3\arrayrulewidth}
 \multicolumn{7}{c}{without refinement} \\
\Xhline{3\arrayrulewidth}
CLIP          & 41.4 & 65.0 & 74.6 & 42.6 & 66.8 & 75.4 \\ 
w/ TS    & 43.2 & 67.8 & 77.2 & 44.2  & 70.5 & 79.1 \\ 
w/ ICS      & 41.0 & 65.2 & 74.5 & 42.1 & 66.9 & 75.0 \\
w/ TS\&ICS   & 43.1 & 67.8 & 77.3 & 44.4 & 70.4  & 79.2  \\ 
\hline
MI-MM   & 37.9 & 61.7 & 72.7 & 38.5 & 63.2 & 73.4 \\ 
w/ TS   & 38.1 & 62.0 & 72.0 & 37.6 & 63.2 & 73.7 \\ 
w/ ICS  &  36.8 & 60.8 & 71.1 & 37.0  & 62.2  & 72.1 \\
w/ TS\&ICS  &  38.1  & 62.8  & 72.7  & 38.1  & 63.2  & 74.2  \\  
\Xhline{3\arrayrulewidth}
 \multicolumn{7}{c}{with refinement} \\
\Xhline{3\arrayrulewidth}
CLIP (paper)          & 50.4 & 74.3 & 80.8 & - & - & - \\ 
CLIP *          & 53.1 & 76.1 & 83.2 & 52.6 & 75.9 & 83.5 \\ 
w/ TS    & 53.1 & 77.1 & 84.4 & 52.6 & 76.3 & 84.5 \\  
w/ ICS    & 52.9 & 75.7 & 83.1 & 52.7 & 76.4 & 83.6 \\
w/ TS\&ICS   & 53.0 & 77.1 & 84.5 & 52.7 & 76.3 & 84.6 \\ 
\hline
MI-MM   & 54.3 & 76.9 & 83.4 & 53.5 & 76.8 & 83.7  \\ 
w/ TS   & 53.8 & 76.6 & 83.4 & 53.3 & 76.8 & 83.5 \\ 
w/ ICS & 54.1  & 76.3 & 82.7 & 54.1 & 77.0  & 83.2 \\  
w/ TS\&ICS  & 53.9   & 76.7  &  83.7 & 52.8  &  76.8 & 83.7   
\\ \hline

\end{tabular}
\caption{ Performance comparison of VAST on YouCook2 dataset (Video retrieval) with modified CLIP and MI-MM losses. TS and ICS temperature (or margin) schedule and
individual cluster shifts, respectively.  "*"- marks our reproduction based on VAST codebase. }
\label{tab:YC2_vast_ablations}
\end{table}

\begin{table*}[h]
\centering
\definecolor{mygreen}{RGB}{0, 150, 0} 
\begin{tabular}{|c|l|c|}
\hline
\textbf{Cluster} & \textbf{Example Text Annotations} & \textbf{Topic} \\
\hline
\makecell[c]{\#1 \\ 106 samples} & \makecell[l]{pour the \colorbox{yellow!40!white}{egg} and add a pinch of salt and red chili powder \\add cooking oil and beat an \colorbox{yellow!40!white}{egg} \\ mix the \colorbox{yellow!40!white}{eggs} with salt and pepper \\ add red pepper flakes to a bowl of \colorbox{yellow!40!white}{eggs} and whisk 
} 
& \makecell[c]{eggs}  \\
\hline
\makecell[c]{\#2 \\ 106 samples} & \makecell[l]{add coconut milk fish sauce and soy sauce into the \colorbox{yellow!40!white}{pan} \\add rice vinegar soy sauce and sriracha to the \colorbox{yellow!40!white}{pan} \\ add oil and green chilies to \colorbox{yellow!40!white}{pan} \\ add garlic thyme bay leaf and tomato paste to the \colorbox{yellow!40!white}{pan} 
}
& \makecell[c]{pan}  \\
\hline
\makecell[c]{\#3 \\ 106 samples } & \makecell[l]{pour the \colorbox{yellow!40!white}{mixture} on hash browns \\ spread the \colorbox{yellow!40!white}{mixture} loosely on the cooking sheet with olive oil and place in the oven \\ \colorbox{yellow!40!white}{mix} the ingredients \\ \colorbox{yellow!40!white}{mix} everything and let it cook 
}  
& \makecell[c]{mixture} \\ \hline
\makecell[c]{\#4 \\ 101 samples} & \makecell[l]{mix flour with the \colorbox{yellow!40!white}{potato} mixture \\  add an egg and farmers cheese to the \colorbox{yellow!40!white}{potatoes} and mash \\ add butter and milk to the pot and mash the \colorbox{yellow!40!white}{potatoes} \\add the mashed \colorbox{yellow!40!white}{potatoes} butter cream and kale to a pot and mix 
} 
& \makecell[c]{ potatoes } \\
\hline
\makecell[c]{\#5 \\ 92 samples} & \makecell[l]{add chopped \colorbox{yellow!40!white}{garlic} chopped ginger and chopped onion \\ add ginger \colorbox{yellow!40!white}{garlic} and onions \\ sprinkle paprika and parsley on top \\ mix red chili sugar \colorbox{yellow!40!white}{garlic} fish sauce and scallians with the radish 
} 
& \makecell[c]{ garlic } \\
\hline
... & ... & ... \\
\hline
\makecell[c]{\#196 \\ 12 samples} & \makecell[l]{mix the hash brown and \colorbox{yellow!40!white}{the sauce} \\ serve the rings with \colorbox{yellow!40!white}{the sauce} \\ serve the beef with \colorbox{yellow!40!white}{mashed potatoes} \\ pour \colorbox{yellow!40!white}{gravy} on the meatloaf 
} 
& \makecell[c]{ sauce } \\
\hline
\makecell[c]{\#197 \\ 12 samples} & \makecell[l]{clean the  \colorbox{yellow!40!white}{liver} and place it on a tray \\ season the \colorbox{yellow!40!white}{liver} \\ slice the \colorbox{yellow!40!white}{liver} \\ place some of the \colorbox{yellow!40!white}{liver} on plastic wrap and roll it 
}
& \makecell[c]{ liver }  \\
\hline
\makecell[c]{\#198 \\ 11 samples} & \makecell[l]{add \colorbox{yellow!40!white}{carrot} and daikon to the bowl and stir \\ drain the \colorbox{yellow!40!white}{carrot} and daikon pickle \\ peel and chop \colorbox{yellow!40!white}{carrot} and daikon into strips and put in a bowl \\ rinse the daikon and \colorbox{yellow!40!white}{carrot} 
} 
& \makecell[c]{ carrot } \\
\hline
 \makecell[c]{\#199 \\ 10 samples} & \makecell[l]{scrape the \colorbox{yellow!40!white}{hummus} onto a dish and top it with some peppers and serve \\ stir the \colorbox{yellow!40!white}{hummus} and add water \\ blend the \colorbox{yellow!40!white}{hummus} \\ add citric acid to the \colorbox{yellow!40!white}{hummus} 
 }
 & \makecell[c]{ hummus } \\
\hline
 \makecell[c]{\#200 \\ 9 samples}  & \makecell[l]{place pieces of bread into the pan and \colorbox{yellow!40!white}{grill} on both sides \\ apply butter on one side of the bread season it with salt and pepper and put bread on \colorbox{yellow!40!white}{grill} \\ combine 2 slices of bread and let it cook on the \colorbox{yellow!40!white}{grill} \\ place the on the grate over the outer ring of charcoal or on the 2nd tier of a gas \colorbox{yellow!40!white}{grill} 
 }
 & \makecell[c]{ grill } \\
\hline
\end{tabular}
\caption{Examples of text annotations (4 per cluster) from 5 biggest clusters (\#1-\#5) and 5 smallest clusters (\#196-\#200), obtained via kMeans clustering on SentenceBERT embeddings of training split of YouCook2 dataset.}
\label{tab:clusters_good_youcook}
\end{table*}

\begin{table*}[h]
\centering
\definecolor{mygreen}{RGB}{0, 150, 0} 
\begin{tabular}{|c|l|c|}
\hline
\textbf{Cluster} & \textbf{Example Text Annotations} & \textbf{Topic} \\
\hline
\makecell[c]{\#1 \\ 69669 samples} & \makecell[l]{scenes of \colorbox{yellow!40!white}{people} working in the office \\ a very long ride with the \colorbox{yellow!40!white}{people} \\ image taken from page of \colorbox{yellow!40!white}{people} \\ index : a list of the topics and \colorbox{yellow!40!white}{people} in the text} & \makecell[c]{people}  \\
\hline
\makecell[c]{\#2 \\ 54898 samples} & \makecell[l]{\colorbox{yellow!40!white}{students} are looking forward to visiting public university \\ \colorbox{yellow!40!white}{students} have fun while they learn about the environment \\ \colorbox{yellow!40!white}{students} prepare for a perennial lesson \\ team up to help keep \colorbox{yellow!40!white}{students} safe} & \makecell[c]{students}  \\
\hline
\makecell[c]{\#3 \\ 38840 samples } & \makecell[l]{\colorbox{yellow!40!white}{football player} was praised after a sensational session between the sticks on tuesday \\ \colorbox{yellow!40!white}{football player} is welcomed back after winning soccer league \\ american \colorbox{yellow!40!white}{football player} will put his body on the line to help his team  \\ \colorbox{yellow!40!white}{football player} is the latest player to be linked with a move away from the club}  & \makecell[c]{football player} 
\\ \hline
\makecell[c]{\#4 \\ 35297 samples} & \makecell[l]{actor attends the european premiere of biographical \colorbox{yellow!40!white}{film} \\  \colorbox{yellow!40!white}{film} director attends the premiere \\ \colorbox{yellow!40!white}{film} director attends premiere held at theater during festival \\ tv writer at the premiere of \colorbox{yellow!40!white}{film}} & \makecell[c]{ film } \\
\hline
\makecell[c]{\#5 \\ 34039 samples} & \makecell[l]{the truth about \colorbox{yellow!40!white}{christmas} vietnam \\ merry \colorbox{yellow!40!white}{christmas} eve to all my wonderful followers ... and to everyone else as well \\ this would be cute to hang on the front door at \colorbox{yellow!40!white}{christmas} time \\ things you have to do this \colorbox{yellow!40!white}{christmas}} & \makecell[c]{ christmas } \\
\hline
\end{tabular}
\caption{Examples of text annotations (4 per cluster) for 5 largest clusters, obtained via K-Means clustering on text embeddings generated using a distilled version of BERT on the training split of CC3M dataset.}
\label{tab:clusters_topcc3m}
\end{table*}

\begin{table*}[h]
\centering
\definecolor{mygreen}{RGB}{0, 150, 0} 
\begin{tabular}{|c|l|c|}
\hline
\textbf{Cluster} & \textbf{Example Text Annotations} & \textbf{Topic} \\
\hline
\makecell[c]{\#196 \\ 3736 samples} & \makecell[l]{\colorbox{yellow!40!white}{man} with glasses looking himself in a mirror isolated \colorbox{yellow!40!white}{on white background} \\ healthy fit young \colorbox{yellow!40!white}{man} measuring his waist with a tape measure to monitor \\ his weight isolated \colorbox{yellow!40!white}{on white background} \\ senior \colorbox{yellow!40!white}{man} and a kid passing a football isolated \colorbox{yellow!40!white}{on white background} \\ portrait of a smiling young \colorbox{yellow!40!white}{man} isolated \colorbox{yellow!40!white}{on white background}\\} & \makecell[c]{ man on a white background } \\
\hline
\makecell[c]{\#197 \\ 3676 samples} & \makecell[l]{\colorbox{yellow!40!white}{silhouette} of young boy playing cricket against a \colorbox{yellow!40!white}{sunset} background \\ \colorbox{yellow!40!white}{trees silhouetted} against a \colorbox{yellow!40!white}{sunset} \\ \colorbox{yellow!40!white}{silhouette} of palm \colorbox{yellow!40!white}{trees} during \colorbox{yellow!40!white}{sunset} at the beach \\ \colorbox{yellow!40!white}{silhouette} of a dead \colorbox{yellow!40!white}{tree} with a \colorbox{yellow!40!white}{sunset} in the background} & \makecell[c]{ silhouette of a tree, sunset}  \\

\hline
\makecell[c]{\#198 \\ 3105 samples} & \makecell[l]{images from the girls \colorbox{yellow!40!white}{basketball} game \\ images from the boys \colorbox{yellow!40!white}{basketball} game \\ images vs. girls \colorbox{yellow!40!white}{basketball} game on thursday , february \\ images from the girls \colorbox{yellow!40!white}{basketball} game in a city} & \makecell[c]{ basketball } \\

\hline
 \makecell[c]{\#199 \\ 3105 samples} & \makecell[l]{beautiful golden pink shining \colorbox{yellow!40!white}{metal} style  \colorbox{yellow!40!white}{uppercase or capital letter} e in a 3d  \\ illustration with a shiny \colorbox{yellow!40!white}{metallic} soft purple red color classic  \colorbox{yellow!40!white}{font} isolated on \\ a white background with clipping path \\ shiny purple \colorbox{yellow!40!white}{metal}  \colorbox{yellow!40!white}{lowercase or small letter} w in a 3d illustration with a rough \\ weathered  \colorbox{yellow!40!white}{metallic} surface texture and classic  \colorbox{yellow!40!white}{font} style isolated on a white \\ background with clipping path \\ shiny \colorbox{yellow!40!white}{metal} silver chrome beveled  \colorbox{yellow!40!white}{lowercase or small letter} n in a 3d illustration \\ with a glossy gray smooth  \colorbox{yellow!40!white}{metallic} surface finish isolated on a white background \\ with clipping path \\ shiny gold metallic  \colorbox{yellow!40!white}{uppercase or capital letter} v in a 3d illustration with a rich \\ golden color and glossy smooth \colorbox{yellow!40!white}{metal} surface finish in a bold  \colorbox{yellow!40!white}{font} isolated on a \\ white background with clipping path \\} & \makecell[c]{ metallic letters } \\
 
\hline
 \makecell[c]{\#200 \\ 2904 samples}  & \makecell[l]{sketches of cupcakes drawn by chalk on a  \colorbox{yellow!40!white}{blackboard}\\ teacher stands at the \colorbox{yellow!40!white}{blackboard} in classroom  \\ little boy to write with chalk on the school  \colorbox{yellow!40!white}{blackboard} \\ white chalk texture vintage stamp with map on a school  \colorbox{yellow!40!white}{blackboard} \\ schoolboy standing in front of a  \colorbox{yellow!40!white}{blackboard}} & \makecell[c]{ blackboard } \\
 
\hline
\end{tabular}
\caption{Examples of text annotations (4 per cluster) for 5 smallest clusters, obtained via K-Means clustering on text embeddings generated using a distilled version of BERT on the training split of CC3M dataset.}
\label{tab:clusters_bottom_cc3m}
\end{table*}

\end{document}